\documentclass[sigconf]{acmart}
\usepackage{multirow}
\AtBeginDocument{%
  \providecommand\BibTeX{{%
    \normalfont B\kern-0.5em{\scshape i\kern-0.25em b}\kern-0.8em\TeX}}}

\setcopyright{acmcopyright}
\copyrightyear{2024}
\acmYear{2024}

\acmConference[London]{London 'The International Conference on Advances in Artificial Intelligence}{October 17--19, 2024}{London, UK}
\acmBooktitle{London: The International Conference on Advances in Artificial Intelligence, October 17--19, 2024, London, UK}
\acmISBN{979-8-4007-1801-4}

\begin{document}

\title{Explainable Artificial Intelligence for Dependent Features: Additive Effects of Collinearity}

\author{Ahmed M Salih}
\email{a.salih@leicester.ac.uk}
\orcid{0000-0002-0871-8282}
\authornotemark[1]
\affiliation{%
  \institution{{Department of Population Health Sciences, University of Leicester}
  \streetaddress{University Rd}
  \city{Leicester}
  \country{United Kingdom}
  \postcode{LE17RH}
}
}


\authornotemark[2]
\affiliation{%
  \institution{{William Harvey Research Institute, Queen Mary University of London}
  \streetaddress{Charterhouse Square}
  \city{London}
  \country{United Kingdom}
  \postcode{EC1M 6BQ}
}
}

\authornotemark[3]
\affiliation{%
  \institution{{Barts Heart Centre, St Bartholomew’s Hospital, Barts Health NHS Trust}
  \streetaddress{West Smithfield}
  \city{London}
  \country{United Kingdom}
  \postcode{EC1A 7BE}
}
}

\authornotemark[4]
\affiliation{%
  \institution{{Department of Computer Science, University of Zakho}
  \streetaddress{Duhok road}
  \city{Zakho}
  \country{Kurdistan of Iraq}
  \postcode{42002}
}
}

\renewcommand{\shortauthors}{Ahmed M Salih}

\begin{abstract}
  Explainable Artificial Intelligence (XAI) emerged to reveal the internal mechanism of machine learning models and how the features affect the prediction outcome. Collinearity is one of the big issues that XAI methods face when identifying the most informative features in the model. Current XAI approaches assume the features in the models are independent and calculate the effect of each feature toward model prediction independently from the rest of the features. However, such assumption is not realistic in real life applications. We propose an Additive Effects of Collinearity (AEC) as a novel XAI method that aim to considers the collinearity issue when it models the effect of each feature in the model on the outcome. AEC is based on the idea of dividing multivariate models into several univariate models in order to examine their impact on each other and consequently on the outcome. The proposed method is implemented using simulated and real data to validate its efficiency comparing with the a state of arts XAI method. The results indicate that AEC is more robust and stable against the impact of  collinearity when it explains AI models compared with the state of arts XAI method.
\end{abstract}






\maketitle

\section{Introduction}
Explainable AI (XAI) has emerged to reveal the internal mechanism of AI models and what are the factors that affect model's decision. List of informative predictors is one of the most common outcome of XAI when it is applied to tabular data. It shows whether a specific feature in the model has effect on the outcome or not and whether the effect is positive or negative.\\
Different methods and approaches were proposed to explain machine learning models locally for a specific instance or globally for all instances in the model. The methods are based on different assumptions and provide different explanation. This includes Shapley Additive Explanations (SHAP)~\cite{NIPS2017} and Local Interpretable Model-agnostic Explanations (LIME)~\cite{lime}. SHAP is based on the game theory to explain a machine learning model where it considers each feature in the model as a player and the outcome as the payoff. LIME generates a local linear model to explain an instance in the model where the coefficient value of the local linear model represents the effect size of the features toward the output.\\
Multicollinearity is one of the big issue XAI faces when it explains a machine learning model. It is a phenomena where the used features in the model are correlated which mean they might affect each other before affecting the outcome of interest. The classic way to interpret the effect of each feature toward the outcome is by reporting the coefficient value of each feature in a multivariate model. However, the interpretation of the coefficient value reflects changing one unit in the feature lead to change in the outcome while holding all other features in the model constant. Such assumption is not realistic when the features are collinear. Most of the current XAI methods assume the features are independent when they explain a model and calculate the effect size of feature toward the model output. Accordingly, the generated explanation does not reflect the interactions between the features.\\
This paper presents an Additive Effects of Collinearity (AEC) which is a novel XAI method that considers the multicollinearity issue among the features. To reveal and calculate the effect of each feature toward the outcome, AEC divides a multivariate model into several univariate models where each feature is considered as dependent and independent feature. The following sections will discuss the related works, the theory behind AEC, its implementation and its validation.
\section{Related-works}
In this section we will discuss some of the current XAI methods that are mostly used in the literature. We will focus on how these methods deal with the collinearity issue and how they explain a model given that the features are collinear. We will consider the XAI methods that provide the list of informative features in the model.
\begin{enumerate}
    \item \textbf{SHAP} is a local, global and agnostic XAI method~\cite{NIPS2017}. It is based on the game theory where each features is considered as a player while the output of the model is considered as the payoff. It considers all collations between the features to calculate a score which represents the impact of the feature toward the model decision. However, SHAP assumes the features in the model are independent when it calculates the impact of a feature. Accordingly, it replaces the conditional distribution by the marginal distribution of the feature of interest. If two features are highly collinear, one of them will get a high score while the second one might get a low score because it does not improve the model performance.
    \item \textbf{LIME} is as its names implies a local model agnostic XAI method ~\cite{lime}. It transfers any model no matter whether it is simple or complex into a local linear model and then reports the coefficient values of the features in the local linear model. The interpretation of the coefficient values represents the effect size (changing in one unit) of one feature on the outcome while holding all other features constant in the model. Such interpretation implies that the features are independent and do not change simultaneously. However, collinearity among the features is very common in most of research area which make such assumption not realistic.
    \item \textbf{Extended Kernel SHAP} is a modified version of Kernel SHAP to explain any model locally~\cite{aas2021explaining}. It estimates the conditional distribution of the features using four approaches based on the distribution of the features. The four approaches are: Gaussian copula distribution, Empirical conditional distribution, Gaussian distribution and the combination of the empirical approach and/or the Gaussian or the Gaussian copula. Although the method fixes one of the important issue, however it is user-dependent as the user has to decide which approach to use. Moreover, it is a local XAI and cannot be extended to a global explanation.
    \item \textbf{Modified Index Position} is a modified version of any XAI that dose not consider collinearity among the features when it calculates the effect toward the outcome~\cite{salih2024characterizing}. It trains a model, apply XAI, remove the top one and then check how many features changed their positions. The method keeps removing and observing how the features change their positions till two features left in the model. The faster the feature reaches the top list, the more important. The authors link the changes in the positions of the features when the top one is removed with the impact of collinearity. Although the method counts on the dependency among the features when it reveal their impact toward the outcome, however it might be computationally expensive.
    \item \textbf{SHAP Cohort Refinement} is another method proposed to modify the SHAP method to consider the collinearity issue~\cite{mase2019explaining}. It makes a new cohort from all samples that is close to the instance need to be explained and then apply SHAP. The method divides the samples into either similar or not to the target using distance-based method. The chosen similarity method affect significantly on whether a specific data point close or not to the target. Consequently, the efficiency of the method depends highly on the similarity method used to identify the cohort.
    \item \textbf{Multi-Collinearity Corrected} is another approach to modify the original version of SHAP to consider the multicollinearity issue~\cite{basu2022multicollinearity}. The method eliminates the impact of features-dependency before calculating the score for each feature. To do so,it adds an adjustment factor to all features that are correlated with the feature of interest and then randomizing the feature of interest. The role of the adjustment factor is to reduce the correlation between the feature of interest with the rest of features. It depends highly on the added factor and to what degree it reduces the impact of correlation.
\end{enumerate}
\section{Proposed method}
\subsection{Theory}
\begin{figure}[H]
    \centering
    \includegraphics[width=6cm, keepaspectratio]{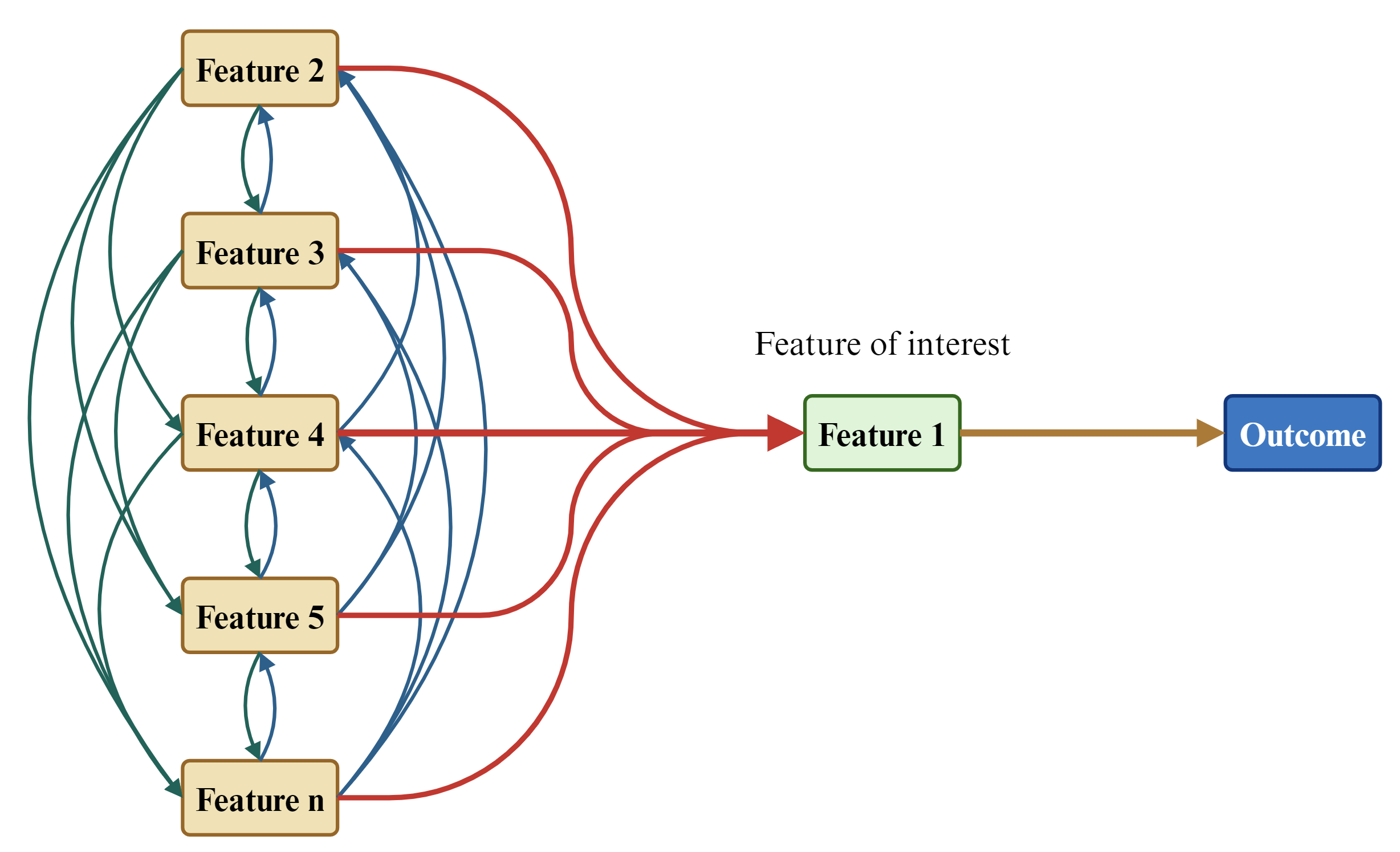}
    \caption{Illustration of the proposed method}
    \label{theory}
\end{figure}

In the presence of collinearity, multivariate models are not able to correctly reveal the impact of one feature on the outcome. It is more reasonable and realistic to dismantled the multivariate model into univariate models to model the impact of the features on each other before their impact on the outcome.\\
Figure~\ref{theory} shows the proposed method where feature 1 is used as the feature of interest just for illustration. It shows that when calculating the effect size of a feature, the effects of all other features  due to multicollinearity among the features will be modeled no matter whether it is big or small.
\noindent
So, let assume we have set of features $X_{k}$ where k is $1,2,3,....,n$ to model an outcome $y$. To model the outcome, multivariate model will be:
\begin{equation}
y \sim \beta_{1} X_{1} + \beta_{2} X_{2} + \beta_{3} X_{3} + \beta_{4} X_{4} + \beta_{5} X_{5} + .......\beta_{n} X_{n}
\end{equation}
Assuming we would like to reveal the impact of $X_{1}$ on the outcome when it is independent from the rest of the features in the model. In this case, we can model the outcome using a univariate model:
\begin{equation}
    y \sim \beta X_{1}
\end{equation}
This is the same as if we add the other features and perform a multivariate model because the interpretation of the coefficient value of $X_{1}$ assumes changing one unit while holding all other features constant in the model. When we used the univariate model, we just assumed the rest of the features are constant.\\ 
In contrary, if the features are collinear, then the impact of $X_{1}$ on the outcome is a combination of the impact of $X_{1}$ on the outcome and the impact the rest of the features on $X_{1}$. In that case, we have to model the impact of the rest of the features on $X_{1}$ and then multiple it with the effect of $X_{1}$ on the outcome. The equation below shows how to model the impact of each feature on $X_{1}$ and modelling the impact of $X_{1}$ on the outcome:
\begin{equation}\label{three}
\textcolor{red}{y \sim \beta X_{1}} * [\textcolor{blue}{X_{1} \sim \beta X_{2}} + \textcolor{teal}{X_{1}\sim  \beta X_{3}} + \textcolor{violet}{X_{1}\sim  \beta X_{4}} + \textcolor{brown}{X_{1} \sim  \beta X_{5}} + ....... + X_{1} \sim \beta X_{n}]
\end{equation}
The multiplication indicates sum the coefficient values from each univariate model and multiply it with the coefficient value of $X_{1}$ toward $y$.
As long we considered the impact of the rest of the features on $X_{1}$ due to the collinearity, we have to consider the impact of the rest of the features on each other because they might be collinear as well. So, equation~\ref{three} can be extended to:
\begin{equation}\label{four}
\begin{split}
\textcolor{blue}{X_{1}\sim \beta X_{2}} = X_{2}  \sim \beta X_{3} + X_{2} \sim \beta X_{4} + X_{2}\sim  \beta X_{5} + .......+ X_{2} \sim \beta X_{n}\\
\textcolor{teal}{X_{1}\sim \beta X_{3}} = X_{3}  \sim \beta X_{2} + X_{3}  \sim \beta X_{4} + X_{3} \sim \beta X_{5}  + .......+ X_{3} \sim \beta X_{n}\\
\textcolor{violet}{X_{1}\sim \beta X_{4}} = X_{4}  \sim \beta X_{2} + X_{4}  \sim \beta X_{3} + X_{4}  \sim \beta X_{5} + ....... + X_{4} \sim \beta X_{n} \\
\textcolor{brown}{X_{1} \sim  \beta X_{5}} = X_{5} \sim \beta X_{2} + X_{5} \sim \beta X_{3} + X_{5} \sim \beta X_{4} +....... + X_{5}  \sim \beta X_{n}
\end{split}
\end{equation}
In equations~\ref{four}, we have modelled the impact of the features on each other considering the same feature might be the dependent or independent feature. It is indeed difficult in many domains to decide whether a specific feature is the dependent or independent one and that is why we considered both cases. Combining the equations in~\ref{three} and \ref{four}, the impact of $X_{1}$ on $y$ considering the collinearity:
\begin{equation}\label{five}
\begin{split}
\textcolor{red}{y \sim \beta X_{1}} * [\textcolor{blue}{X_{1} \sim \beta X_{2}} + [X_{2}  \sim \beta X_{3} + X_{2} \sim \beta X_{4}  + X_{2}\sim  \beta X_{5} \\ + .......+ X_{2} \sim \beta X_{n}] + \\ \textcolor{teal}{X_{1}\sim  \beta X_{3}}  + [X_{3}  \sim \beta X_{2} + X_{3}  \sim \beta X_{4} + X_{3} \sim \beta X_{5} \\  + .......+ X_{3} \sim \beta X_{n}] +  \\ \textcolor{violet}{X_{1}\sim  \beta X_{4}} + [X_{4}  \sim \beta X_{2} + X_{4}  \sim \beta X_{3} + X_{4}  \sim \beta X_{5} \\ + ....... + X_{4} \sim \beta X_{n}] + \\ \textcolor{brown}{X_{1} \sim  \beta X_{5}} +[ X_{5} \sim \beta X_{2} + X_{5} \sim \beta X_{3} + X_{5} \sim \beta X_{4} \\ + ....... + X_{5}  \sim \beta X_{n} ] + \\.......... + X_{1} \sim \beta X_{n}]
\end{split}
\end{equation}
Finally, The equations in~\ref{five} can be combined into one single equation:
\begin{equation}
\textcolor{red}{y \sim \beta X_{1}} * \sum_{z=1}^{n}\sum_{i=1}^{n}X_{z} \sim \beta X_{i} 
\end{equation}
where $i \neq  1$ and $z\neq i$.\\
Consequently, from the equations above, the impact of any feature on the outcome considering the collinearity among the features is:
\begin{equation}
\textcolor{red}{y \sim \beta X_{j}} * \sum_{z=1}^{n}\sum_{i=1}^{n}X_{z} \sim \beta X_{i} 
\end{equation}
where $i \neq  j$ and $z\neq i$ and $j$ is the feature of interest, 

\subsection{Properties}
\begin{enumerate}
    \item \textbf{Inclusively}: AEC is inclusive in terms of considering the effects of the features on each other and on the outcome. When univariate models are applied, we consider both cases where the same feature might be the independent or dependent variable because it is difficult to decide which one of them affect the other.
    
    \item \textbf{Additive effects}: Unlike coefficient value in multivariate models, AEC has the property that it calculates the additive effect among all the features when it calculates the effect size of a feature toward the outcome. It might be argued that this increases the number or running models and consequently the computation time. In addition, it might be suggested that it would be better to not model the association between two features if they are not highly collinear and consequently decrease the computation time. However, even a small association should be considered because it might lead to a big effect when considering the accumulative effects.
    
    \item \textbf{Meaningful value}: The proposed method calculates the effect size of each feature in its own unit. The effect size can be interpreted as the changing one unit in the feature of interest and the other features in the model toward the model outcome.

    \item \textbf{Flexibility}: In the implementation section we applied a linear regression and a logistic regression model to predict a continues variable and to perform classification. However, these two models can be replaced to other AI models which they might provide better accuracy depending on the domain.
\end{enumerate}

\section{Implementation}
To implement the proposed method, we used two kinds of datasets. Firstly, we simulated datasets to perform regression and classification. Secondly, we used real datasets to perform similar tasks. We considered SHAP as a state of arts XAI method to compare our method with because it is one of the most widely used XAI method in the literature~\cite{holzinger2022explainable}. Moreover, it provides list of informative features globally which fits nicely with AEC. The code of the implementation for a regression model can be find at~\url{https://github.com/amaa11/Additive-Effects-of-Collinearity}.
\subsection{Simulated data}
\begin{figure}[H]
    \includegraphics[width=6.5cm]{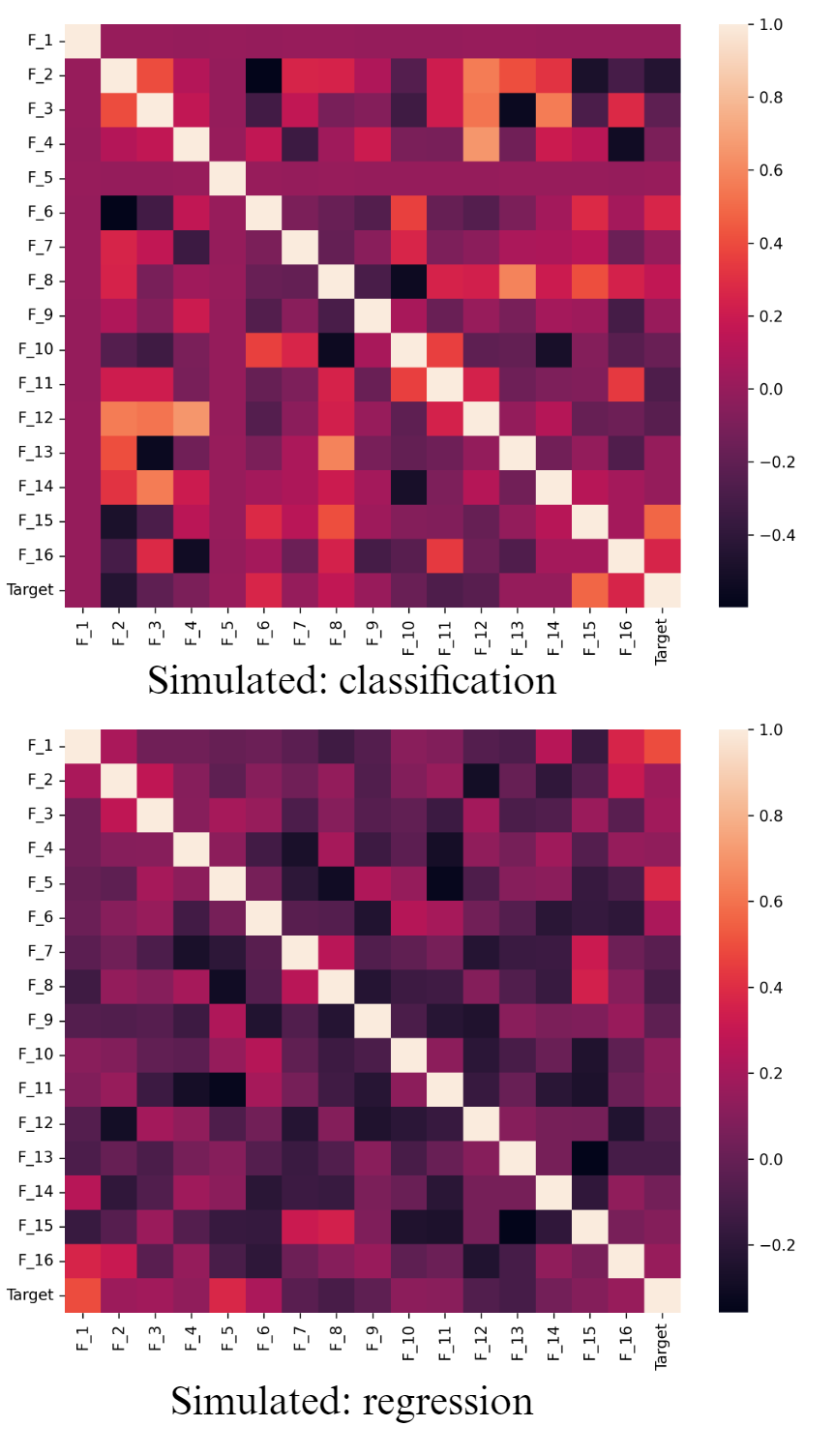}%
    \caption{Correlation heatmap of the simulated datasets}\label{re_sim}
\end{figure}

We simulated two sets of data to perform classification and regression using a linear model. The reason of using simulated data is to generate collinear features and examine the effectiveness of the proposed method to explain the model in the presence of such phenomena.
\begin{enumerate}
    \item \textbf{Classification Model}: We used the Sklearn library in Python with \textit{make\_classification} to generate data for a classification model. The generated features were 16, with 9 informative, 5 redundant and 2 noise. The features were generated for 100,000 samples. Logistic regression was used to perform classification using the generated features. The proposed method generated the list of informative features in 40 seconds using a PC with Intel(R) Core(TM) i5-6500 CPU, RAM: 8.00 GB and Windows 10 64-bit operating system.
    \item \textbf{Regression model}: Sklearn library in Python was used with \textit{make\_regression} to generate data for a regression model. The generated features were 16 features, with 7 informative for 150,000 samples. Multivariate linear regression model was used to predict the outcome using the generated features. The proposed method generated the list of significant features in 57 seconds using the same PC mentioned before.
\end{enumerate}
Figures~\ref{re_sim} shows the correlation heatmap of the generated features for the classification and regression models. The figure shows that there is collinearity among the features which ultimately will affect the explanation model. 

\noindent
Table~\ref{Corre_regr} shows the list of informative features generated by for AEC and SHAP in both classification and regression model. The tables shows that there is a variation in the list between both methods even in the top once. For example, in the regression model the feature \textit{F15} was the third one in SHAP method while it was the eleventh in the AEC. In the classification task, \textit{F2} is the top one on the list in SHAP method while it is the fifth one in AEC method.
\begin{table}[H]
\centering
\caption{List of the informative predictors for AEC and SHAP.}
\begin{tabular}{|cc|cc|}
\hline
\multicolumn{2}{|c|}{\textbf{Regression}}          & \multicolumn{2}{c|}{\textbf{Classification}}      \\ \hline
\multicolumn{1}{|c|}{\textbf{AEC}} & \textbf{SHAP} & \multicolumn{1}{c|}{\textbf{AEC}} & \textbf{SHAP} \\ \hline
\multicolumn{1}{|c|}{F1}           & F5            & \multicolumn{1}{c|}{F15}          & F2            \\ \hline
\multicolumn{1}{|c|}{F5}           & F1            & \multicolumn{1}{c|}{F6}           & F11           \\ \hline
\multicolumn{1}{|c|}{F3}           & F15           & \multicolumn{1}{c|}{F16}          & F15           \\ \hline
\multicolumn{1}{|c|}{F16}          & F11           & \multicolumn{1}{c|}{F10}          & F16           \\ \hline
\multicolumn{1}{|c|}{F6}           & F4            & \multicolumn{1}{c|}{F2}           & F9            \\ \hline
\multicolumn{1}{|c|}{F2}           & F6            & \multicolumn{1}{c|}{F11}          & F12           \\ \hline
\multicolumn{1}{|c|}{F4}           & F13           & \multicolumn{1}{c|}{F12}          & F10           \\ \hline
\multicolumn{1}{|c|}{F10}          & F7            & \multicolumn{1}{c|}{F3}           & F4            \\ \hline
\multicolumn{1}{|c|}{F8}           & F10           & \multicolumn{1}{c|}{F8}           & F7            \\ \hline
\multicolumn{1}{|c|}{F13}          & F8            & \multicolumn{1}{c|}{F4}           & F6            \\ \hline
\multicolumn{1}{|c|}{F15}          & F16           & \multicolumn{1}{c|}{F9}           & F13           \\ \hline
\multicolumn{1}{|c|}{F11}          & F3            & \multicolumn{1}{c|}{F1}           & F8            \\ \hline
\multicolumn{1}{|c|}{F12}          & F14           & \multicolumn{1}{c|}{F5}           & F14           \\ \hline
\multicolumn{1}{|c|}{F14}          & F2            & \multicolumn{1}{c|}{F7}           & F3            \\ \hline
\multicolumn{1}{|c|}{F7}           & F9            & \multicolumn{1}{c|}{F13}          & F1            \\ \hline
\multicolumn{1}{|c|}{F9}           & F12           & \multicolumn{1}{c|}{F14}          & F5            \\ \hline
\end{tabular}\label{Corre_regr}
\end{table}
\subsection{Real data}
\begin{figure}[H]
    \includegraphics[width=7cm]{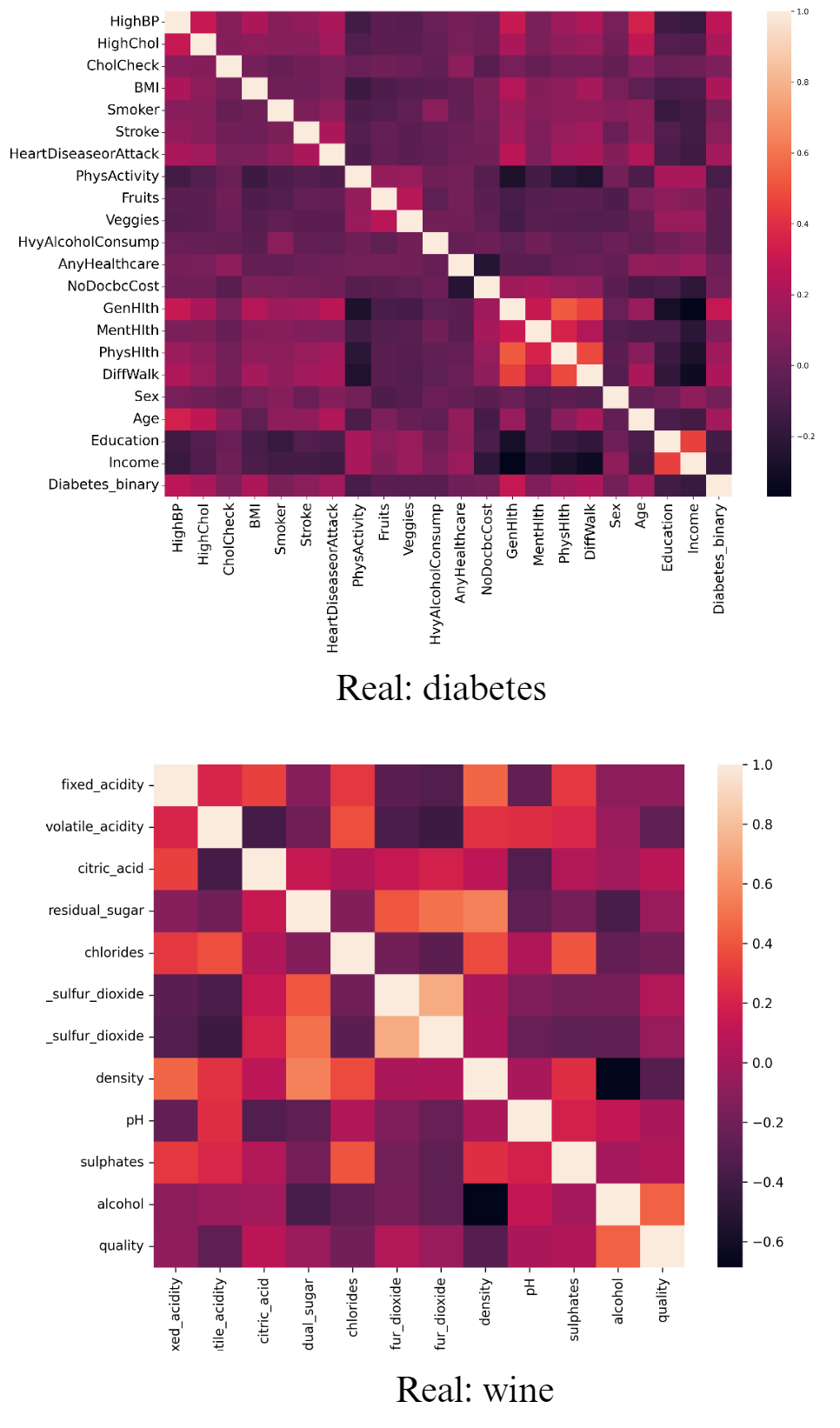}%
  \caption{Correlation heatmap of the real datasets}\label{corr_real_wine}
\end{figure}

\begin{enumerate}
    \item \textbf{The Diabetes Health Indicators Dataset}: The first real dataset was from Diabetes Health Indicators Dataset downloaded from UCI Machine Learning Repository~\cite{burrows2017incidence}. The dataset consists of 21 features involving both continues and categorical variables to classify 253,680 samples into free of diabetes and with diabetes. Logistic regression was used to perform binary classification to classify subjects into control or with diabetes. The proposed method last 3.35 minutes to run and calculates the effect size for each feature using the same PC mentioned above.
    \item \textbf{The Wine Quality Dataset}: The second real dataset was the Wine Quality indicator which comprises 12 features to predict the quality of the wine for 4,898 instances. The datasets was downloaded from UCI Machine Learning Repository~\cite{cortez2009modeling}. Multivariate linear regression model was used to predict the quality of the wine because it ranges from 0 to 10. The proposed method last 4 seconds to run and calculates the effect size for each feature using the same PC mentioned above.
\end{enumerate}
Figure~\ref{corr_real_wine} shows the correlation heatmap among the features for both the Diabetes and Wine quality datasets.

\noindent
Table~\ref{Wine_dia} shows the list of informative features for both Wine quality and Diabetes datasets using AEC and SHAP methods. The tables shows that there is a significant variation in the list between AEC and SHAP. For instance, \textit{density} is the top feature to predict the quality of wine using the AEC method while it was the fourth one using SHAP method. Similarly, the \textit{CholCheck} is on the top of the list to predict samples with diabetes using AEC method while it was the tenth using SHAP method. In the validation section we explained which one of the two lists is more robust against the collinearity issue and provide a more realistic list.
\begin{table}[t]
\scriptsize
\centering
\caption{List of the informative predictors for AEC and SHAP using Wine quality and Diabetes datasets}
\begin{tabular}{|cc|cc|}
\hline
\multicolumn{2}{|c|}{\textbf{Wine quality}}                           & \multicolumn{2}{c|}{\textbf{Diabetes}}                           \\ \hline
\multicolumn{1}{|c|}{\textbf{AEC}}           & \textbf{SHAP}          & \multicolumn{1}{c|}{\textbf{AEC}}         & \textbf{SHAP}        \\ \hline
\multicolumn{1}{|c|}{density}                & alcohol                & \multicolumn{1}{c|}{CholCheck}            & GenHlth              \\ \hline
\multicolumn{1}{|c|}{chlorides}              & residual\_sugar        & \multicolumn{1}{c|}{HighBP}               & HighBP               \\ \hline
\multicolumn{1}{|c|}{volatile\_acidity}      & volatile\_acidity      & \multicolumn{1}{c|}{HeartDiseaseorAttack} & HighChol             \\ \hline
\multicolumn{1}{|c|}{citric\_acid}           & density                & \multicolumn{1}{c|}{HighChol}             & BMI                  \\ \hline
\multicolumn{1}{|c|}{alcohol}                & total\_sulfur\_dioxide & \multicolumn{1}{c|}{HvyAlcoholConsump}    & Age                  \\ \hline
\multicolumn{1}{|c|}{sulphates}              & sulphates              & \multicolumn{1}{c|}{DiffWalk}             & Income               \\ \hline
\multicolumn{1}{|c|}{pH}                     & free\_sulfur\_dioxide  & \multicolumn{1}{c|}{Stroke}               & HvyAlcoholConsump    \\ \hline
\multicolumn{1}{|c|}{fixed\_acidity}         & fixed\_acidity         & \multicolumn{1}{c|}{PhysActivity}         & Sex                  \\ \hline
\multicolumn{1}{|c|}{residual\_sugar}        & pH                     & \multicolumn{1}{c|}{GenHlth}              & Fruits               \\ \hline
\multicolumn{1}{|c|}{free\_sulfur\_dioxide}  & citric\_acid           & \multicolumn{1}{c|}{Veggies}              & CholCheck            \\ \hline
\multicolumn{1}{|c|}{total\_sulfur\_dioxide} & chlorides              & \multicolumn{1}{c|}{Education}            & AnyHealthcare        \\ \hline
\multicolumn{1}{|c|}{\multirow{10}{*}{}}     & \multirow{10}{*}{}     & \multicolumn{1}{c|}{Smoker}               & NoDocbcCost          \\ \cline{3-4} 
\multicolumn{1}{|c|}{}                       &                        & \multicolumn{1}{c|}{NoDocbcCost}          & PhysHlth             \\ \cline{3-4} 
\multicolumn{1}{|c|}{}                       &                        & \multicolumn{1}{c|}{Fruits}               & Education            \\ \cline{3-4} 
\multicolumn{1}{|c|}{}                       &                        & \multicolumn{1}{c|}{AnyHealthcare}        & DiffWalk             \\ \cline{3-4} 
\multicolumn{1}{|c|}{}                       &                        & \multicolumn{1}{c|}{Income}               & MentHlth             \\ \cline{3-4} 
\multicolumn{1}{|c|}{}                       &                        & \multicolumn{1}{c|}{Age}                  & HeartDiseaseorAttack \\ \cline{3-4} 
\multicolumn{1}{|c|}{}                       &                        & \multicolumn{1}{c|}{Sex}                  & Smoker               \\ \cline{3-4} 
\multicolumn{1}{|c|}{}                       &                        & \multicolumn{1}{c|}{BMI}                  & Veggies              \\ \cline{3-4} 
\multicolumn{1}{|c|}{}                       &                        & \multicolumn{1}{c|}{PhysHlth}             & PhysActivity         \\ \cline{3-4} 
\multicolumn{1}{|c|}{}                       &                        & \multicolumn{1}{c|}{MentHlth}             & Stroke               \\ \hline
\end{tabular}\label{Wine_dia}
\end{table}

\section{Validation}
To validate the proposed method and compare it with SHAP, we applied Normalized Movement Rate (NMR)~\cite{salih2022investigating} method to assess the impact of collinearity on the quality of the outcomes of AEC and SHAP. NMR measures numerically the impact of collinearity on the list of informative features generated by any XAI method. It removes the top feature generated by XAI method iteratively and then observe how the order of the list of the informative features is changed which is the impact of collinearity. NMR value ranges between 0 and 1; more close to zero means the model is more robust against the collinearity while more close to one indicates the list of informative features generated by the XAI method is not robust against collinearity and biased.\\
We applied NMR on the outcomes of both AEC and SHAP to compare their performance in terms of the robustness of the list against collinearity with the two simulated and real datasets. Table~\ref{NMR} shows the NMR values for both AEC and SHAP in the four datasets. The tables hows that in the four cases the NMR value in the AEC is smaller than the one generated by SHAP. This indicates that the list generated by AEC experienced less dis-order when the top one was removed and consequently more robust in the presence of collinearity compared to SHAP.
\begin{table}[H]
\small
\caption{Normalized movement rate value for AEC and SHAP using the four datasets}
    \centering
    \begin{tabular}{|l|c|c|}
    \hline
        \textbf{Dataset} & \textbf{AEC}  & \textbf{SHAP} \\ \hline
       Simulated Classification dataset  & 0.225  & 0.361 \\ \hline
       Simulated Regression dataset  & 0.052  & 0.325 \\ \hline
       Diabetes Health Indicators dataset  &0.107  & 0.179\\ \hline
       Wine Quality dataset & 0.201 & 0.337 \\ \hline
    \end{tabular}
    \label{NMR}
\end{table}
\section{Discussion}
XAI methods are used more frequently than before because it has become one of the main element of AI systems nowadays. However, less attention have been given to the one of the main issue that affect the whole explanation. Collinearity occurs usually in real life data and applications and need to be considered to provided a more precise explanation. Without considering such significant issue, the explanation does not really reflect the internal mechanism of the AI models.
In this work, we presented a novel XAI method "AEC" that considers and models the association among the features and their impact on model prediction. AEC has been implemented with four collinear datasets, among them two real data while the other two are simulated. In the four datasets, the method has proved to provide a more robust and stable explanation comparing to the state of arts method. Moreover, the method has several favorable properties including easy to model and understand. Such method is in favor of the end users where the majority of them from non AI background. 


\bibliographystyle{ACM-Reference-Format}
\bibliography{sample-base}


\begin{thebibliography}{10}


\ifx \showCODEN    \undefined \def \showCODEN     #1{\unskip}     \fi
\ifx \showDOI      \undefined \def \showDOI       #1{#1}\fi
\ifx \showISBNx    \undefined \def \showISBNx     #1{\unskip}     \fi
\ifx \showISBNxiii \undefined \def \showISBNxiii  #1{\unskip}     \fi
\ifx \showISSN     \undefined \def \showISSN      #1{\unskip}     \fi
\ifx \showLCCN     \undefined \def \showLCCN      #1{\unskip}     \fi
\ifx \shownote     \undefined \def \shownote      #1{#1}          \fi
\ifx \showarticletitle \undefined \def \showarticletitle #1{#1}   \fi
\ifx \showURL      \undefined \def \showURL       {\relax}        \fi
\providecommand\bibfield[2]{#2}
\providecommand\bibinfo[2]{#2}
\providecommand\natexlab[1]{#1}
\providecommand\showeprint[2][]{arXiv:#2}

\bibitem[\protect\citeauthoryear{Aas, Jullum, and L{\o}land}{Aas
  et~al\mbox{.}}{2021}]%
        {aas2021explaining}
\bibfield{author}{\bibinfo{person}{Kjersti Aas}, \bibinfo{person}{Martin
  Jullum}, {and} \bibinfo{person}{Anders L{\o}land}.}
  \bibinfo{year}{2021}\natexlab{}.
\newblock \showarticletitle{Explaining individual predictions when features are
  dependent: More accurate approximations to Shapley values}.
\newblock \bibinfo{journal}{\emph{Artificial Intelligence}}
  \bibinfo{volume}{298} (\bibinfo{year}{2021}), \bibinfo{pages}{103502}.
\newblock


\bibitem[\protect\citeauthoryear{Basu and Maji}{Basu and Maji}{2022}]%
        {basu2022multicollinearity}
\bibfield{author}{\bibinfo{person}{Indranil Basu} {and}
  \bibinfo{person}{Subhadip Maji}.} \bibinfo{year}{2022}\natexlab{}.
\newblock \showarticletitle{Multicollinearity correction and combined feature
  effect in shapley values}. In \bibinfo{booktitle}{\emph{Australasian Joint
  Conference on Artificial Intelligence}}. Springer, \bibinfo{pages}{79--90}.
\newblock


\bibitem[\protect\citeauthoryear{Burrows}{Burrows}{2017}]%
        {burrows2017incidence}
\bibfield{author}{\bibinfo{person}{Nilka~Rios Burrows}.}
  \bibinfo{year}{2017}\natexlab{}.
\newblock \showarticletitle{Incidence of end-stage renal disease attributed to
  diabetes among persons with diagnosed diabetes—United States and Puerto
  Rico, 2000--2014}.
\newblock \bibinfo{journal}{\emph{MMWR. Morbidity and mortality weekly report}}
   \bibinfo{volume}{66} (\bibinfo{year}{2017}).
\newblock


\bibitem[\protect\citeauthoryear{Cortez, Cerdeira, Almeida, Matos, and
  Reis}{Cortez et~al\mbox{.}}{2009}]%
        {cortez2009modeling}
\bibfield{author}{\bibinfo{person}{Paulo Cortez}, \bibinfo{person}{Ant{\'o}nio
  Cerdeira}, \bibinfo{person}{Fernando Almeida}, \bibinfo{person}{Telmo Matos},
  {and} \bibinfo{person}{Jos{\'e} Reis}.} \bibinfo{year}{2009}\natexlab{}.
\newblock \showarticletitle{Modeling wine preferences by data mining from
  physicochemical properties}.
\newblock \bibinfo{journal}{\emph{Decision support systems}}
  \bibinfo{volume}{47}, \bibinfo{number}{4} (\bibinfo{year}{2009}),
  \bibinfo{pages}{547--553}.
\newblock


\bibitem[\protect\citeauthoryear{Holzinger, Saranti, Molnar, Biecek, and
  Samek}{Holzinger et~al\mbox{.}}{2022}]%
        {holzinger2022explainable}
\bibfield{author}{\bibinfo{person}{Andreas Holzinger}, \bibinfo{person}{Anna
  Saranti}, \bibinfo{person}{Christoph Molnar}, \bibinfo{person}{Przemyslaw
  Biecek}, {and} \bibinfo{person}{Wojciech Samek}.}
  \bibinfo{year}{2022}\natexlab{}.
\newblock \showarticletitle{Explainable AI methods-a brief overview}. In
  \bibinfo{booktitle}{\emph{International workshop on extending explainable AI
  beyond deep models and classifiers}}. Springer, \bibinfo{pages}{13--38}.
\newblock


\bibitem[\protect\citeauthoryear{Lundberg and Lee}{Lundberg and Lee}{2017}]%
        {NIPS2017}
\bibfield{author}{\bibinfo{person}{Scott~M Lundberg} {and}
  \bibinfo{person}{Su-In Lee}.} \bibinfo{year}{2017}\natexlab{}.
\newblock \showarticletitle{A Unified Approach to Interpreting Model
  Predictions}.
\newblock In \bibinfo{booktitle}{\emph{Advances in Neural Information
  Processing Systems 30}}, \bibfield{editor}{\bibinfo{person}{I.~Guyon},
  \bibinfo{person}{U.~V. Luxburg}, \bibinfo{person}{S.~Bengio},
  \bibinfo{person}{H.~Wallach}, \bibinfo{person}{R.~Fergus},
  \bibinfo{person}{S.~Vishwanathan}, {and} \bibinfo{person}{R.~Garnett}}
  (Eds.). \bibinfo{publisher}{Curran Associates, Inc.},
  \bibinfo{pages}{4765--4774}.
\newblock
\urldef\tempurl%
\url{http://papers.nips.cc/paper/7062-a-unified-approach-to-interpreting-model-predictions.pdf}
\showURL{%
\tempurl}


\bibitem[\protect\citeauthoryear{Mase, Owen, and Seiler}{Mase
  et~al\mbox{.}}{2019}]%
        {mase2019explaining}
\bibfield{author}{\bibinfo{person}{Masayoshi Mase}, \bibinfo{person}{Art~B
  Owen}, {and} \bibinfo{person}{Benjamin Seiler}.}
  \bibinfo{year}{2019}\natexlab{}.
\newblock \showarticletitle{Explaining black box decisions by shapley cohort
  refinement}.
\newblock \bibinfo{journal}{\emph{arXiv preprint arXiv:1911.00467}}
  (\bibinfo{year}{2019}).
\newblock


\bibitem[\protect\citeauthoryear{Ribeiro, Singh, and Guestrin}{Ribeiro
  et~al\mbox{.}}{2016}]%
        {lime}
\bibfield{author}{\bibinfo{person}{Marco~Tulio Ribeiro},
  \bibinfo{person}{Sameer Singh}, {and} \bibinfo{person}{Carlos Guestrin}.}
  \bibinfo{year}{2016}\natexlab{}.
\newblock \showarticletitle{"Why Should {I} Trust You?": Explaining the
  Predictions of Any Classifier}. In \bibinfo{booktitle}{\emph{Proceedings of
  the 22nd {ACM} {SIGKDD} International Conference on Knowledge Discovery and
  Data Mining, San Francisco, CA, USA, August 13-17, 2016}}.
  \bibinfo{publisher}{ACM}, \bibinfo{pages}{1135--1144}.
\newblock


\bibitem[\protect\citeauthoryear{Salih, Galazzo, Cruciani, Brusini, and
  Radeva}{Salih et~al\mbox{.}}{2022}]%
        {salih2022investigating}
\bibfield{author}{\bibinfo{person}{Ahmed Salih},
  \bibinfo{person}{Ilaria~Boscolo Galazzo}, \bibinfo{person}{Federica
  Cruciani}, \bibinfo{person}{Lorenza Brusini}, {and} \bibinfo{person}{Petia
  Radeva}.} \bibinfo{year}{2022}\natexlab{}.
\newblock \showarticletitle{Investigating explainable artificial intelligence
  for mri-based classification of dementia: a new stability criterion for
  explainable methods}. In \bibinfo{booktitle}{\emph{2022 IEEE International
  Conference on Image Processing (ICIP)}}. IEEE, \bibinfo{publisher}{IEEE},
  \bibinfo{pages}{4003--4007}.
\newblock


\bibitem[\protect\citeauthoryear{Salih, Galazzo, Raisi-Estabragh, Petersen,
  Menegaz, and Radeva}{Salih et~al\mbox{.}}{2024}]%
        {salih2024characterizing}
\bibfield{author}{\bibinfo{person}{Ahmed~M Salih},
  \bibinfo{person}{Ilaria~Boscolo Galazzo}, \bibinfo{person}{Zahra
  Raisi-Estabragh}, \bibinfo{person}{Steffen~E Petersen},
  \bibinfo{person}{Gloria Menegaz}, {and} \bibinfo{person}{Petia Radeva}.}
  \bibinfo{year}{2024}\natexlab{}.
\newblock \showarticletitle{Characterizing the Contribution of Dependent
  Features in XAI Methods}.
\newblock \bibinfo{journal}{\emph{IEEE Journal of Biomedical and Health
  Informatics}} (\bibinfo{year}{2024}).
\newblock


\end{thebibliography}
\end{document}